\documentclass[conference]{IEEEtran}
\IEEEoverridecommandlockouts
\usepackage{cite}
\usepackage{amsmath,amssymb,amsfonts}
\usepackage{algorithmic}
\usepackage{graphicx}
\usepackage{textcomp}
\usepackage{xcolor}
\usepackage{multirow}
\usepackage{booktabs}
\usepackage{tabularx}
\usepackage{array}
\usepackage{float}
\usepackage{placeins}

\def\BibTeX{{\rm B\kern-.05em{\sc i\kern-.025em b}\kern-.08em
    T\kern-.1667em\lower.7ex\hbox{E}\kern-.125emX}}
\begin{document}

\title{MCST-Mamba: Multivariate Mamba-Based Model for Traffic Prediction}

\author{
\IEEEauthorblockN{Mohamed Hamad}
\IEEEauthorblockA{Department of Electrical Engineering,\\ College of Engineering\\
Qatar University\\
Doha, Qatar\\
m.bashir.hamad@qu.edu.qa}
\and
\IEEEauthorblockN{Mohamed Mabrok}
\IEEEauthorblockA{Department of Mathematics and Physics,\\ College of Arts and Sciences\\
Qatar University\\
Doha, Qatar\\
m.a.mabrok@qu.edu.qa}
\and
\IEEEauthorblockN{Nizar Zorba}
\IEEEauthorblockA{Department of Electrical Engineering,\\ College of Engineering\\
Qatar University\\
Doha, Qatar\\
nizarz@qu.edu.qa}
}

\maketitle

\begin{abstract}
Accurate traffic prediction plays a vital role in intelligent transportation systems by enabling efficient routing, congestion mitigation, and proactive traffic control. However, forecasting is challenging due to the combined effects of dynamic road conditions, varying traffic patterns across different locations, and external influences such as weather and accidents. Traffic data often consists of several interrelated measurements—such as speed, flow and occupancy—yet many deep‐learning approaches either predict only one of these variables or require a separate model for each. This limits their ability to capture joint patterns across channels. To address this, we introduce the Multi-Channel Spatio-Temporal (MCST) Mamba model, a forecasting framework built on the Mamba selective state-space architecture that natively handles multivariate inputs and simultaneously models all traffic features. The proposed MCST-Mamba model integrates adaptive spatio-temporal embeddings and separates the modeling of temporal sequences and spatial sensor interactions into two dedicated Mamba blocks, improving representation learning. Unlike prior methods that evaluate on a single channel, we assess MCST-Mamba across all traffic features at once, aligning more closely with how congestion arises in practice. Our results show that MCST-Mamba achieves strong predictive performance with a lower parameter count compared to baseline models.
\end{abstract}

\section{Introduction}
\label{sec:introduction}

Accurate traffic forecasting is critical for intelligent transportation systems (ITS), as it enables proactive traffic management, improves road safety, and reduces congestion through better route planning and resource allocation \cite{yin2022deep}. Despite the huge progress, both from research and implementation, traffic forecasting still presents several key challenges.  
First, traffic time series depend on a multitude of external factors—such as special events, accidents, and weather—that are difficult to model explicitly \cite{nizae2020GCN}.  
Second, modern sensor networks involve hundreds or thousands of detectors whose interactions are governed by complex road layouts \cite{nizar2018iDriveSense}; where spatially adjacent sensors may behave quite differently, making it hard to infer connectivity.
Third, traffic data from typical sensor networks is often intrinsically multi-channelled (speed, occupancy, density, etc.), and traffic congestion is the result of the complex relationship between these traffic network features.

Recent deep-learning approaches have used different learning methods to advance the state of the art \cite{taskGCN} \cite{mabrok2024uav}. However, many popular architectures that use Transformers or Graph Convolutional Networks (GCN) forecast only a single variable (typically speed) per sensor, even when multivariate measurements are available \cite{liu2023staeformer}. When additional channels are included, they often require training separate univariate models for each channel—either in parallel (increasing parameter count and compute) or sequentially (increasing total training time)—which limits their ability to learn inter-channel dependencies in a unified framework.

Recently, the Mamba architecture was introduced \cite{gu2023mamba}, a selective state space model (SSM) designed for efficient sequence modeling. Mamba captures long-range dependencies in sequential data while maintaining near-linear computational complexity, making it particularly suitable for time series forecasting tasks. Crucially, unlike many existing models that require separate processing for each data channel, Mamba processes the full multivariate input as a single sequence, enabling it to capture inter-channel dependencies naturally and efficiently within a unified architecture. Mamba has demonstrated superior performance over transformer-based baselines in several domains \cite{Li2024Semi, yang2025vivim}, and initial studies confirm its potential for traffic forecasting, reporting gains in both accuracy and efficiency \cite{wang2024mamba_tsf, shao2024stmambasync}.

In this paper, we introduce \textit{MCST-Mamba}, which addresses the above challenges using four approaches:  
\begin{enumerate}
  \item We exploit Mamba's intrinsic multi-channel interface to simulataneously train our model on all variables, increasing computational efficiency and inter-channel dependency extraction.
  \item We propose STAEformer-style adaptive embedding \cite{liu2023staeformer} for Mamba, which effectively captures complex spatio-temporal relations and chronological information in traffic time series.
  \item We apply Mamba as two independent blocks: one for spatial dependencies and one for temporal dependencies, each fed with appropriately formatted inputs and combined.  
  \item Unlike typical traffic forecasting approaches that report accuracy on a single variable (e.g., speed), we evaluate MCST-Mamba across all available channels simultaneously. This better reflects real-world congestion dynamics.
\end{enumerate}
These strategies align our model architecture more closely with the spatio-temporal, multivariate nature of traffic data and deliver strong performance with reduced number of parameters.

\section{Related Work}
\subsection{Mamba for Sequence Modeling}

Mamba \cite{gu2023mamba} is a recent state space model (SSM) designed for long-range sequence modeling with high computational efficiency. It builds on structured state space models, which describe an input sequence \(\mathbf{u}(t)\) using continuous-time latent dynamics:
\begin{subequations} \label{eq:ssm_continuous}
\begin{align}
\dot{\mathbf{x}}(t) &= \mathbf{A}\,\mathbf{x}(t) + \mathbf{B}\,\mathbf{u}(t), \label{eq:ssm_continuous_state}\\
\mathbf{y}(t)       &= \mathbf{C}\,\mathbf{x}(t) + \mathbf{D}\,\mathbf{u}(t), \label{eq:ssm_continuous_output}
\end{align}
\end{subequations}
so that discretizing over a fixed interval \(\Delta\) gives a linear recurrence:
\begin{subequations} \label{eq:ssm_discrete}
\begin{align}
\mathbf{x}_k &= \mathbf{\bar{A}}\,\mathbf{x}_{k-1} + \mathbf{\bar{B}}\,\mathbf{u}_k, \label{eq:ssm_discrete_state}\\
\mathbf{y}_k &= \mathbf{C}\,\mathbf{x}_k + \mathbf{D}\,\mathbf{u}_k, \label{eq:ssm_discrete_output}
\end{align}
\end{subequations}

Mamba extends this formulation by making the transition matrices input-dependent:  
\begin{equation} \label{eq:mamba_input_dependent}
\mathbf{A}_k = f_A(\mathbf{u}_k),\quad \mathbf{B}_k = f_B(\mathbf{u}_k),\quad \mathbf{C}_k = f_C(\mathbf{u}_k),
\end{equation}
allowing the model to dynamically adjust its information propagation based on the current input. This selective updating mechanism enables Mamba to effectively model nonlinear temporal dynamics \cite{gu2023mamba}.

Crucially for traffic forecasting, Mamba accepts multivariate input directly in the form \(\mathbf{U} \in \mathbb{R}^{b \times \ell \times d}\), where \(b\) is the batch size, \(\ell\) the sequence length, and \(d\) the number of features.
Unlike current traffic prediction models implemented using Transformers or GCN that often treat each traffic variable (e.g., speed, flow, occupancy) independently, Mamba processes all channels jointly within a unified sequence, capturing inter-channel correlations without additional architectural complexity.

Furthermore, Mamba achieves efficient sequence modeling through a parallel prefix-sum ("scan") algorithm, yielding \(\mathcal{O}(n)\) work and \(\mathcal{O}(\log n)\) depth complexity. This makes it particularly suitable for real-time or resource-constrained forecasting scenarios like traffic management, where long-range patterns and low latency are essential.

Recent studies have shown that Mamba outperforms Transformers in various domains, including time series forecasting \cite{wang2024mamba_tsf, shao2024stmambasync}, motivating its use in our model for spatio-temporal multivariate traffic prediction.

\subsection{Embeddings Strategies}

Spatio‐temporal adaptive embedding was first introduced in \cite{liu2023staeformer} to enhance vanilla Transformers, demonstrating that a learnable tensor \(\mathbf{E}_a\) can effectively capture intrinsic spatio‐temporal relations and chronological information in traffic time series \cite{liu2023staeformer}. In their model, feature embedding \(\mathbf{E}_f\) projects raw traffic series into a higher‐dimensional space via a linear layer, while periodicity embedding \(\mathbf{E}_p\) incorporates day‐of‐week and time‐of‐day signals through learned lookup tables. The core adaptive embedding \(\mathbf{E}_a\) is a learnable tensor that jointly encodes spatial node identities and temporal context, enabling the model to infer latent dependencies without explicit graphs.

In contrast to STAEformer's use of Transformer layers, our work adopts the same spatio‐temporal adaptive embedding \(\mathbf{E}_a\) but feeds the concatenated hidden spatio-temporal representation \(\mathbf{Z}\), defined as: 
\begin{equation} \label{eq:staeformer_embedding_concat}
\mathbf{Z} = \bigl[\mathbf{E}_f \,\|\, \mathbf{E}_p \,\|\, \mathbf{E}_a\bigr]
\end{equation}
into Mamba's selective SSM blocks.

\section{Methodology}
\label{sec:methodology}
We formulate traffic speed prediction as a spatiotemporal forecasting problem. The raw input data is structured as a 4-dimensional tensor \(\mathbf{X} \in \mathbb{R}^{m \times t_{in} \times n \times f}\), where:
$m$ is the batch size, $t_{in}$ is the length of the input time window (historical observations), $n$ is the number of sensors/nodes in the traffic network, and $f$ is the feature dimension, representing one or more traffic measurements.

The feature dimension $f$, in the used multi-channel datasets, consists of three variables that characterize traffic conditions at 5-minute intervals:
\begin{itemize}
    \item Average Speed (v): The mean speed of vehicles detected during the 5-minute interval, measured in kilometers (or miles) per hour. This can be measured directly by dual-loop detectors or derived from single-loop data.
    \item Traffic Flow (q): The total number of vehicles passing a sensor location in the 5-minute intervals. This is typically obtained by summing raw vehicle counts collected at shorter intervals (e.g., 30 seconds).
    \item Average Occupancy (o): The fraction of time within the 5-minute interval that the sensor was occupied by vehicles. It is calculated as \(o = (\text{Total occupied time} / \text{Interval length}) \times 100\%\).
\end{itemize}
Using these three variables together ($f=3$) provides a more comprehensive representation of traffic states compared to single-channel approaches that often focus only on speed ($f=1$). As highlighted in Section~\ref{sec:introduction}, traffic congestion is a result of the complex interplay between flow, speed, and occupancy. Additionally, the MCST-Mamba architecture is specifically designed to process multivariate inputs natively, making it well-suited to leverage the richness of these multi-channel datasets.

\begin{figure*}[t]
    \centering
    \includegraphics[width=1\linewidth]{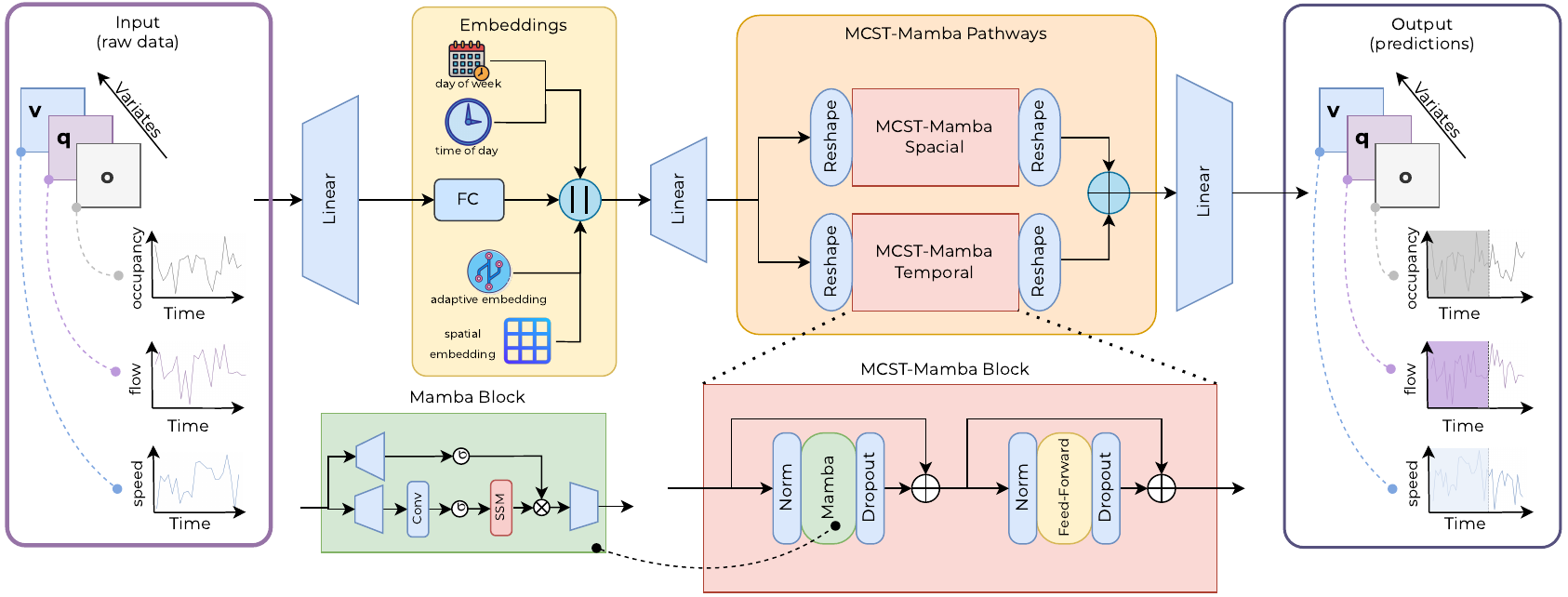}
    \caption{MCST-Mamba Architecture}
    \label{fig:arch}
\end{figure*}

The prediction task can be mathematically formulated as:
\begin{equation} \label{eq:pred_task}
[\mathbf{X}_{t+1}, \mathbf{X}_{t+2}, ..., \mathbf{X}_{t+t_{out}}] = f([\mathbf{X}_{t-t_{in}+1}, \mathbf{X}_{t-t_{in}+2}, ..., \mathbf{X}_t])
\end{equation}
where $f$ is the prediction function that maps historical traffic measurements to future traffic states across all nodes in the network. Figure \ref{fig:arch} provides an overview of the architecture of the MCST-Mamba model.

\subsection{Embeddings and Data Preparation}

In our model, we transform the raw traffic data into a rich representation space through multiple embedding strategies adopted from the STAEformer architecture. The embedding process captures temporal patterns and spatial characteristics, providing the model with comprehensive context.

Given the input tensor \(\mathbf{X} \in \mathbb{R}^{m \times t_{\text{in}} \times n \times f}\), we construct the final multi-component embedding \(\mathbf{Z}\) as follows:
\begin{equation} \label{eq:embedding_concat}
\mathbf{Z} = \text{Concat}(\mathbf{E}_{f}, \mathbf{E}_{p}, \mathbf{E}_{\text{spatial}}, \mathbf{E}_{a})
\end{equation}
where \(Concat\) merges the embeddings along their last dimension. The periodicity embedding \(\mathbf{E}_p\) itself is formed by concatenating time-of-day and day-of-week embeddings: \(\mathbf{E}_p = \text{Concat}(\mathbf{E}_{\text{tod}}, \mathbf{E}_{\text{dow}})\). Note that our \(\mathbf{Z}\) explicitly includes the spatial embedding \(\mathbf{E}_{\text{spatial}}\), extending the formulation presented in STAEformer. Each component is detailed below:

\subsubsection{Feature Embedding (\(\mathbf{E}_f\))} 

We project the primary traffic features (flow, speed, occupancy) in \(\mathbf{X}\) through a linear transformation to obtain \(\mathbf{E}_{f}\), a 24-dimensional embedding. This allows the model to better capture relationships between different traffic metrics.

\subsubsection{Temporal Embeddings (\(\mathbf{E}_{tod}, \mathbf{E}_{dow}\) forming \(\mathbf{E}_p\))}

To recognize recurring patterns, we incorporate time-of-day (\(\mathbf{E}_{\text{tod}}\)) and day-of-week (\(\mathbf{E}_{\text{dow}}\)) embeddings. For time-of-day, each 5-minute interval is mapped to one of 288 slots and embedded as \(\mathbf{E}_{\text{tod}}\). For day-of-week, each position is encoded with a categorical embedding \(\mathbf{E}_{\text{dow}}\). These temporal embeddings help distinguish rush hours, weekdays, and weekends. They are concatenated to form the periodicity embedding \(\mathbf{E}_p\).

\subsubsection{Spatial Embedding (\(\mathbf{E}_{\text{spatial}}\))} 

Each node is associated with a unique 16-dimensional embedding vector. These vectors are stored row-wise in the learnable spatial embedding matrix \(\mathbf{E}_{\text{spatial}} \in \mathbb{R}^{n \times d_{\text{spatial}}}\), differentiating locations and their characteristics. This helps capture unique location properties.

\subsubsection{Adaptive Embedding (\(\mathbf{E}_a\))}

We enhance the representation with \(\mathbf{E}_{a}\), an 80-dimensional position-aware embedding capturing node- and timestep-specific information. This enables adaptation to different spatiotemporal contexts.

The combined embedding tensor \(\mathbf{Z}\) is then projected to the model's internal dimension using a weight \(\mathbf{W}_{\text{proj}}\) and a bias \(\mathbf{b}_{\text{proj}}\):
\begin{equation} \label{eq:embedding_projected}
\mathbf{E}_{\text{proj}} = \mathbf{W}_{\text{proj}} \mathbf{Z} + \mathbf{b}_{\text{proj}}
\end{equation}
where, $\mathbf{W}_{\text{proj}} \in \mathbb{R}^{D_{\text{model}} \times D_{\text{mamba}}}$ and $\mathbf{b}_{\text{proj}} \in \mathbb{R}^{D_{\text{mamba}}}$, with $D_{\text{mamba}}$ being the internal dimension used by the Mamba blocks. This comprehensive embedding strategy allows the model to effectively capture various aspects of traffic patterns, including temporal dependencies, spatial characteristics, and position-specific information, providing a rich foundation for the subsequent Mamba processing.

\subsection{Mamba Block Implementation}

After the embedding process, our model employs a dual-pathway architecture that separately processes temporal and spatial dimensions through specialized Mamba blocks. This design enables the model to efficiently capture both long-range temporal dependencies and complex spatial relationships without explicit graph modeling.

\subsubsection{State Space Model Blocks}
The core computational unit in our architecture is the MCST-Mamba block, which implements the Mamba SSM enhanced with normalization and residual connections. Each block first applies layer normalization to stabilize learning, then processes the data through the Mamba SSM component, and finally adds the result back to the input through a residual connection. This is followed by a feed-forward network that further transforms the representation while maintaining the original information through another residual connection.

\subsubsection{Temporal Processing Pathway}

Traffic data is naturally 4D, with shape \([m, n, t, f]\), but Mamba operates on 3D input of shape \([b, l, d]\), requiring us to reshape the data accordingly.

To model temporal dynamics, we treat each node's multivariate time series as an independent sequence. We first permute and reshape the input tensor \(\mathbf{X} \in \mathbb{R}^{m \times n \times t_{in} \times f}\) as follows:
\begin{equation} \label{eq:reshape_temporal}
\begin{split}
&\mathbf{X}_{\text{temporal}} \\ &= \text{Reshape}\Big(  \text{Permute}(\mathbf{X},\ (1,\ 0,\ 2,\ 3)),  [n,\ m \cdot t_{in},\ f] \Big)
\end{split}
\end{equation}

The \(\text{Permute}\) operation reorders the dimensions of the input tensor according to the specified sequence, while \(\text{Reshape}\) reorganizes the data into a new structure with different dimensions while preserving the total number of elements. Together, these operations transform the 4D traffic data into a 3D format suitable for Mamba processing, where each node's temporal evolution becomes single sequence, with all channels processed jointly to capture inter-variable temporal dynamics.

\subsubsection{Spatial Processing Pathway}

To model spatial dependencies, we treat each timestep as a spatial snapshot of the entire network. We begin with the same input tensor \(\mathbf{X} \in \mathbb{R}^{m \times n \times t_{in} 
\times f}\) and permute and reshape it as follows:
\begin{equation} \label{eq:reshape_spatial}
\begin{split}
&\mathbf{X}_{\text{spatial}} \\&=( \text{Reshape}\Big(\text{Permute}(\mathbf{X},\ (2,\ 0,\ 1,\ 3)),\ [t_{in},\ m \cdot n,\ f]\Big)
\end{split}
\end{equation}

This yields \(\mathbf{X}_{\text{spatial}} \in \mathbb{R}^{t_{in} \times (m \cdot n) \times f}\), allowing Mamba t o process the multivariate signals across all nodes simultaneously at each timestep. This configuration enables the model to implicitly learn spatial dependencies, such as congestion propagation or inter-node influence, without relying on a predefined graph structure.

\subsubsection{Adaptive Feature Integration}
The key insight of our dual-pathway architecture is that temporal and spatial patterns in traffic data can be processed independently and then integrated. After processing through their respective Mamba blocks, we denote the outputs as \(\mathbf{Y}_{\text{temporal}}\) and \(\mathbf{Y}_{\text{spatial}}\). We return both outputs to the original shape and combine them using a weighted sum with learnable scalar weights:
\begin{equation} \label{eq:combined_output}
\mathbf{Y}_{\text{combined}} = \mathbf{Y}_{\text{temporal}} w_{\text{temporal}} + \mathbf{Y}_{\text{spatial}} w_{\text{spatial}}
\end{equation}
\noindent where $w_{\text{temporal}}$ and $w_{\text{spatial}}$ are learnable scalar parameters that determine the relative importance of each pathway for the final prediction.
This integration mechanism allows the model to adaptively emphasize temporal or spatial patterns depending on their predictive utility for different nodes and time periods. The combined representation \(\mathbf{Y}_{\text{combined}}\) then undergoes normalization and projection to produce the final prediction.

\subsubsection{Learning Process Intuition}
The learning process in our model is guided by three key intuitions:
\begin{enumerate}
    \item 
Decomposition Principle: Traffic patterns can be effectively decomposed into temporal evolution (how traffic at a location changes over time) and spatial configuration (how traffic distributes across the network at a given time).
    \item 
State-based Modeling: Both temporal and spatial dynamics can be modeled as state transitions in a continuous system, which allows for efficient computation and effective capturing of long-range dependencies.
    \item 
Adaptive Integration: The relative importance of temporal and spatial patterns varies across different scenarios and should be learned from data rather than fixed a priori.
\end{enumerate}
This design enables our model to efficiently process traffic data without requiring explicit graph structure information, while still capturing complex spatiotemporal dependencies through the state space formulation.

\section{Experiments}
We evaluate our proposed MCST-Mamba model on multiple traffic datasets and compare it with state-of-the-art baseline models.

\subsection{Experimental Setup}
All experiments were conducted on a workstation with an Intel Xeon E5-2690 v4 CPU, 220 GB of RAM, and 2x Tesla V100-PCIE GPUs with 16 GB of memory, running Ubuntu 24.04.2 LTS. Our model was implemented using PyTorch 2.0.0 and the Mamba library. The code is available at \verb|https://github.com/mombash/MCST-Mamba|

\subsection{Datasets}
We evaluate our model on two widely-used, multi-channel traffic prediction benchmarks from the California Performance Measurement System (PeMS): PEMS-D4 and PEMS-D8 \cite{guo2019astgcn}.
The selection of these datasets is motivated by several key considerations. First, the PEMS datasets are widely recognized benchmarks in traffic prediction research, allowing for direct comparison with existing methods. Second, both datasets are multi-channel, containing speed, flow, and occupancy measurements, which provides a richer representation of traffic conditions compared to single-channel datasets. This multi-channel setup enables the model to learn more comprehensive traffic patterns by considering different aspects of traffic state simultaneously. Table \ref{tab:dataset} summarizes the key statistics of these datasets.

\begin{table}[ht]
\centering
\caption{Dataset Statistics}
\label{tab:dataset}
\small
\setlength{\tabcolsep}{3pt} 
\renewcommand{\arraystretch}{1.2} 
\begin{tabularx}{\columnwidth}{lccccc}
\toprule
\textbf{Dataset} & 
\textbf{Sensors} & 
\textbf{Time Steps} & 
\textbf{\shortstack{Time\\Range}} & 
\textbf{\shortstack{Time\\Interval}} & 
\textbf{\shortstack{Features}} \\
\midrule
PEMS-D4 & 307 & 16,992 & 1/18--2/18 & 5 min & \shortstack{Speed,\\Flow,\\Occupancy} \\
PEMS-D8 & 170 & 17,856 & 7/16--8/16 & 5 min & \shortstack{Speed,\\Flow,\\Occupancy} \\
\bottomrule
\end{tabularx}
\end{table}
\small

\begin{table*}[t]
\centering
\caption{Performance comparison on PEMS04 and PEMS08 datasets.}
\label{tab:overall}
\small
\setlength{\tabcolsep}{4pt} 
\renewcommand{\arraystretch}{1.2} 
\begin{tabularx}{\textwidth}{l|*{2}{>{\centering\arraybackslash}X|>{\centering\arraybackslash}X|>{\centering\arraybackslash}X}}
\hline
\multirow{2}{*}{\textbf{Model}} & \multicolumn{3}{c|}{\textbf{PEMS04}} & \multicolumn{3}{c}{\textbf{PEMS08}} \\
\cline{2-7}
& MAE & RMSE & MAPE & MAE & RMSE & MAPE \\
\hline
HI & 42.35 & 61.66 & 29.92\% & 36.66 & 50.45 & 21.63\% \\
\hline
GWNet & 18.53 & 29.92 & 12.89\% & 14.40 & 23.39 & 9.21\% \\
DCRNN & 19.63 & 31.26 & 13.59\% & 15.22 & 24.17 & 10.21\% \\
AGCRN & 19.38 & 31.25 & 13.40\% & 15.32 & 24.41 & 10.03\% \\
STGCN & 19.57 & 31.38 & 13.44\% & 16.08 & 25.39 & 10.60\% \\
GTS & 20.96 & 32.95 & 14.66\% & 16.49 & 26.08 & 10.54\% \\
MTGNN & 19.17 & 31.70 & 13.37\% & 15.18 & 24.24 & 10.20\% \\
\hline
STNorm & 18.96 & 30.98 & 12.69\% & 15.41 & 24.77 & 9.76\% \\
GMAN & 19.14 & 31.60 & 13.19\% & 15.31 & 24.92 & 10.13\% \\
\hline
PDFormer & 18.36 & 30.03 & 12.00\% & 13.58 & 23.41 & 9.05\% \\
STID & 18.38 & 29.95 & 12.04\% & 14.21 & 23.28 & 9.27\% \\
STAEformer & 18.22 & 30.18 & \underline{11.98\%} & 13.46 & 23.25 & \underline{8.88\%} \\
\hline
ST-Mamba & \underline{18.19} & 30.17 & \textbf{11.88\%} & 13.40 & 23.20 & 9.00\% \\
ST-MambaSync & 18.20 & \underline{29.85} & 12.00\% & \underline{13.30} & \underline{23.14} & \textbf{8.80\%} \\
MCST-Mamba (Ours) & \textbf{7.88} & \textbf{20.97} & 16.05\% & \textbf{6.54} & \textbf{16.43} & 12.06\% \\
\hline
\end{tabularx}
\end{table*}

Following standard practice, we use 70\% of the data for training, 10\% for validation, and 20\% for testing. 
The data is normalized using z-score normalization based on the training set statistics.
For all datasets, we use 12 historical time steps (1 hour) as input to simultaneously predict the next 12 time steps of traffic speed, flow and occupancy.

\subsection{Baseline Models}
We compare our MCST-Mamba model with several state-of-the-art models across five categories:

\begin{itemize}
    \item \textbf{Traditional Statistical Models}: Approaches like Historical Average (HI) that rely on historical data patterns without complex learning mechanisms (Kashyap \emph{et al.}, 2021)\cite{Kashyap2021Traffic}.
    
    \item \textbf{Graph Convolutional Methods}: Models that leverage graph structures to capture spatial dependencies in traffic networks, including Graph WaveNet (GWNet) and DCRNN. These methods use different mechanisms to model spatial relationships between traffic sensors (Medina-Salgado \emph{et al.}, 2022)\cite{MedinaSalgado2022Urban}.
    
    \item \textbf{General Spatio-Temporal Models}: Architectures like STNorm and GMAN that combine various techniques to jointly model spatial and temporal dependencies without explicitly using graph neural networks or transformers.
    
    \item \textbf{Transformer-based Methods}: Models that adapt the transformer architecture for traffic prediction, such as PDFormer and STAEformer, leveraging self-attention mechanisms to capture long-range dependencies.
    
    \item \textbf{State Space Models}: Recent approaches like ST-Mamba that apply selective state space models to traffic prediction, offering efficient alternatives to attention-based models while maintaining competitive performance.
\end{itemize}

\subsection{Evaluation Metrics}

We use three widely-adopted metrics to evaluate model performance:

\begin{itemize}
    \item Mean Absolute Error: $\text{MAE} = \frac{1}{N_{\text{samples}}} \sum_{i=1}^{N_{\text{samples}}} |y_i - \hat{y}_i|$
    \item Root Mean Squared Error: $\text{RMSE} = \sqrt{\frac{1}{N_{\text{samples}}} \sum_{i=1}^{N_{\text{samples}}} (y_i - \hat{y}_i)^2}$
    \item Mean Absolute Percentage Error: $\text{MAPE} = \frac{1}{N_{\text{samples}}} \sum_{i=1}^{N_{\text{samples}}} \left| \frac{y_i - \hat{y}_i}{y_i} \right| \times 100\%$
\end{itemize}
where, $y_i$ is the ground truth value, $\hat{y}_i$ is the predicted scalar value, and $N_{\text{samples}}$ is the total number of samples. Lower values indicate better performance for all metrics.

\subsection{Implementation Details}
We implement our MCST-Mamba model with the following hyperparameters:
\begin{itemize}
\item Embedding dimensions: $D_{\text{input}}=24$, $D_{\text{tod}}=24$, $D_{\text{dow}}=24$, $D_{\text{spatial}}=16$, $D_{\text{adaptive}}=80$
\item Mamba block dimensions: $D_{\text{mamba}}=96$
\item Number of Mamba blocks: 1 per pathway (temporal and spatial)
\item State dimension: $N=32$
\item Expansion factor: 2
\item Dropout rate: 0.1
\end{itemize}
We train our model using the Adam optimizer with an initial learning rate of 0.001, with a cosine annealing learning rate scheduler. We use early stopping with a patience of 15 epochs. Batch size is set to 64 for most datasets, with adjustments for larger datasets. We train all models with the Mean Squared Error (MSE) loss function.
\subsection{Results and Analysis}

\subsubsection{Overall Performance}
Table \ref{tab:overall} presents the overall performance of various models on the PEMS04 and PEMS08 datasets, with prediction horizons of 12 steps (1 hour). The baseline results are reported in the STAEformer paper \cite{liu2023staeformer}, where the metrics reflect performance on a single predicted channel (speed). In contrast, the metrics for MCST-Mamba represent the aggregated error across all three predicted channels (speed, flow, occupancy), reflecting a more comprehensive evaluation.

Our MCST-Mamba model demonstrates state-of-the-art performance in terms of Mean Absolute Error (MAE) and Root Mean Squared Error (RMSE) on both the PEMS04 and PEMS08 datasets. The model significantly outperforms all baseline methods on these metrics. For instance, on PEMS04, MCST-Mamba achieves an MAE of 7.88 and RMSE of 20.97, representing reductions of approximately 56.7\% and 29.7\% respectively, compared to the best-performing baselines (ST-Mamba for MAE and ST-MambaSync for RMSE). Similar substantial improvements are observed on PEMS08, with MAE reduced by 50.8\% (6.54 vs. 13.30) and RMSE by 29.0\% (16.43 vs. 23.14) compared to ST-MambaSync.

Interestingly, while MCST-Mamba excels in MAE and RMSE, its Mean Absolute Percentage Error (MAPE) is higher than several recent baselines like ST-Mamba, ST-MambaSync, and STAEformer on both datasets. This discrepancy suggests that while the model minimizes absolute errors effectively, it might produce larger relative errors, potentially on predictions where the ground truth values are small, a known sensitivity of the MAPE metric.

\subsubsection{Efficiency Analysis}
We also evaluate the parameter efficiency of our model. Table \ref{tab:efficiency} shows the number of parameters for our model and several baselines.

\begin{table}[H]
\caption{Model size comparison in number of parameters}
\centering
\begin{tabular}{lc}
\hline
\textbf{Model} & \textbf{Parameters} \\
\hline
DCRNN & 0.38M \\
GWNET & 0.28M \\
GTS & 20.17M \\
MTGNN & 0.43M \\
STAEformer & 1.2M \\
MCST-Mamba (Ours) & 0.49M \\
\hline
\end{tabular}
\label{tab:efficiency}
\end{table}

As shown in Table \ref{tab:efficiency}, our MCST-Mamba model contains 0.49M parameters. This size is comparable to several graph-based methods like MTGNN (0.43M) and DCRNN (0.38M), and significantly more compact than the Transformer-based STAEformer (1.2M) and the very large GTS (20.17M).

A critical point is that the listed baseline models are typically evaluated on a single output channel (e.g., speed prediction). In contrast, our MCST-Mamba model processes and predicts all three input channels (speed, flow, occupancy) concurrently using this parameter budget. Therefore, considering the amount of information processed per parameter, MCST-Mamba achieves substantially higher parameter efficiency. It effectively handles three times the feature dimensionality compared to single-channel models of a similar parameter count, showcasing the efficiency of our proposed MCST-Mamba approach.

\subsection{Discussion}
Our experimental results demonstrate the effectiveness of the MCST-Mamba architecture for traffic prediction. The dual pathway design, which processes temporal and spatial dimensions separately, enables the model to capture complex spatiotemporal dependencies more effectively than existing approaches. The integration of adaptive embeddings further enhances the model's ability to learn meaningful representations from traffic data.

The consistent performance improvements across different datasets suggest that our approach is robust and generalizable. The efficiency analysis also confirms that the MCST-Mamba model strikes a good balance between prediction accuracy and computational cost, making it suitable for real-world traffic prediction applications where both performance and efficiency are important.

\section{Conclusion}
In this paper, we introduced MCST-Mamba, a novel multivariate traffic forecasting model based on the selective state space Mamba architecture. We addressed the inherent challenges of traffic prediction, namely the complex spatio-temporal dependencies and the multi-channel nature of traffic data (speed, flow, occupancy), which are often handled inadequately by models that focus on univariate prediction or require separate processing for each channel.

Our core contribution lies in leveraging the Mamba architecture's native ability to process multivariate sequences efficiently. MCST-Mamba incorporates adaptive spatio-temporal embeddings inspired by STAEformer and employs a dual-pathway design where separate Mamba blocks model temporal evolution and spatial interactions. These pathways are adaptively integrated, allowing the model to learn the relative importance of temporal and spatial patterns. We evaluated MCST-Mamba by predicting all traffic channels simultaneously, providing a more holistic assessment compared to typical single-channel evaluations.

Experimental results on the PEMS-D4 and PEMS-D8 benchmark datasets demonstrated that MCST-Mamba achieves state-of-the-art performance on multiple metrics. Furthermore, MCST-Mamba maintains competitive parameter efficiency, comparable to several single-channel models despite handling three times the feature dimensions.

Future work will focus on several promising directions. We plan to investigate advanced model fusion techniques to enhance the integration between the temporal and spatial Mamba pathways. Incorporating additional relevant data channels, such as weather conditions, traffic incidents, and public events, could further improve prediction accuracy by providing richer contextual information. Additionally, we plan to explore the use of comprehensive model fusion techniques for fusing the outputs of the temporal and spatial Mamba pathways to further improve prediction accuracy.

Overall, MCST-Mamba presents an effective and efficient approach for multivariate spatio-temporal traffic forecasting, offering significant potential for improving intelligent transportation systems.


\end{document}